\documentclass[10pt,journal,compsoc]{IEEEtran}
\ifCLASSOPTIONcompsoc
  \usepackage[nocompress]{cite}
\else
  \usepackage{cite}
\fi
\ifCLASSINFOpdf
   \usepackage[pdftex]{graphicx}
\else
\fi
\usepackage{amsmath}
\usepackage{array}
\usepackage{url}
\usepackage{color}%For color
\usepackage{algpseudocode}
\usepackage{algorithm}%For algorithms
\usepackage{amsfonts}%For math symbols

\usepackage{float} % BEWARE: load before caption

\usepackage{caption}
  \DeclareCaptionType{copyrightbox}
  \usepackage{subcaption}
  
\usepackage{booktabs} %For tables
\usepackage{hhline}
\usepackage{tikz}
\usepackage{lipsum}

\usepackage{multirow}

\begin{document}
%
% paper title
% Titles are generally capitalized except for words such as a, an, and, as,
% at, but, by, for, in, nor, of, on, or, the, to and up, which are usually
% not capitalized unless they are the first or last word of the title.
% Linebreaks \\ can be used within to get better formatting as desired.
% Do not put math or special symbols in the title.
%\title{The Design of Efficient Mitchell's Approximate Log Multiplier for Convolutional Neural Networks}
\title{The Effects of Approximate Multiplication on Convolutional Neural Networks}
%
% author names and IEEE memberships
% note positions of commas and nonbreaking spaces ( ~ ) LaTeX will not break
% a structure at a ~ so this keeps an author's name from being broken across
% two lines.
% use \thanks{} to gain access to the first footnote area
% a separate \thanks must be used for each paragraph as LaTeX2e's \thanks
% was not built to handle multiple paragraphs
%
%
%\IEEEcompsocitemizethanks is a special \thanks that produces the bulleted
% lists the Computer Society journals use for "first footnote" author
% affiliations. Use \IEEEcompsocthanksitem which works much like \item
% for each affiliation group. When not in compsoc mode,
% \IEEEcompsocitemizethanks becomes like \thanks and
% \IEEEcompsocthanksitem becomes a line break with idention. This
% facilitates dual compilation, although admittedly the differences in the
% desired content of \author between the different types of papers makes a
% one-size-fits-all approach a daunting prospect. For instance, compsoc 
% journal papers have the author affiliations above the "Manuscript
% received ..."  text while in non-compsoc journals this is reversed. Sigh.

\author{
        Min Soo Kim,
        Alberto A. Del Barrio,~\IEEEmembership{Senior Member,~IEEE},
        HyunJin Kim,
        Nader Bagherzadeh,~\IEEEmembership{Fellow,~IEEE}% <-this % stops a space
}

\IEEEtitleabstractindextext{%
\begin{abstract}
This paper analyzes the effects of approximate multiplication when performing inferences on deep convolutional neural networks (CNNs). The approximate multiplication can reduce the cost of the underlying circuits so that CNN inferences can be performed more efficiently in hardware accelerators. The study identifies the critical factors in the convolution, fully-connected, and batch normalization layers that allow more accurate CNN predictions despite the errors from approximate multiplication. The same factors also provide an arithmetic explanation of why bfloat16 multiplication performs well on CNNs. The experiments are performed with recognized network architectures to show that the approximate multipliers can produce predictions that are nearly as accurate as the FP32 references, without additional training. For example, the ResNet and Inception-v4 models with Mitch-$w$6 multiplication produces Top-5 errors that are within 0.2\% compared to the FP32 references. A brief cost comparison of Mitch-$w$6 against bfloat16 is presented where a MAC operation saves up to 80\% of energy compared to the bfloat16 arithmetic. The most far-reaching contribution of this paper is the analytical justification that multiplications can be approximated while additions need to be exact in CNN MAC operations.
\end{abstract}

% Note that keywords are not normally used for peerreview papers.
% \begin{IEEEkeywords}
% Approximate Multiplier, Convolutional Neural Networks, Logarithm, Power Reduction, MNIST, CIFAR-10, ImageNet
% \end{IEEEkeywords}}

% Note that keywords are not normally used for peerreview papers.
\begin{IEEEkeywords}
Machine learning , Computer vision,  Object recognition, Arithmetic and logic units, Low-power design 
\end{IEEEkeywords}}

The manuscript has been accepted for publication in the IEEE Transactions on Emerging Topics in Computing.

\hfill

IEEE Copyright Notice

\hfill

© 2021 IEEE.  Personal use of this material is permitted.  Permission from IEEE must be obtained for all other uses, in any current or future media, including reprinting/republishing this material for advertising or promotional purposes, creating new collective works, for resale or redistribution to servers or lists, or reuse of any copyrighted component of this work in other works.
\newpage

% make the title area
\maketitle

%\copyrightnotice
% To allow for easy dual compilation without having to reenter the
% abstract/keywords data, the \IEEEtitleabstractindextext text will
% not be used in maketitle, but will appear (i.e., to be "transported")
% here as \IEEEdisplaynontitleabstractindextext when the compsoc 
% or transmag modes are not selected <OR> if conference mode is selected 
% - because all conference papers position the abstract like regular
% papers do.
\IEEEdisplaynontitleabstractindextext
% \IEEEdisplaynontitleabstractindextext has no effect when using
% compsoc or transmag under a non-conference mode.

% For peer review papers, you can put extra information on the cover
% page as needed:
% \ifCLASSOPTIONpeerreview
% \begin{center} \bfseries EDICS Category: 3-BBND \end{center}
% \fi
%
% For peerreview papers, this IEEEtran command inserts a page break and
% creates the second title. It will be ignored for other modes.
\IEEEpeerreviewmaketitle

\IEEEraisesectionheading{\section{Introduction}\label{sec:introduction}}
% Computer Society journal (but not conference!) papers do something unusual
% with the very first section heading (almost always called "Introduction").
% They place it ABOVE the main text! IEEEtran.cls does not automatically do
% this for you, but you can achieve this effect with the provided
% \IEEEraisesectionheading{} command. Note the need to keep any \label that
% is to refer to the section immediately after \section in the above as
% \IEEEraisesectionheading puts \section within a raised box.

% The very first letter is a 2 line initial drop letter followed
% by the rest of the first word in caps (small caps for compsoc).
% 
% form to use if the first word consists of a single letter:
% \IEEEPARstart{A}{demo} file is ....
% 
% form to use if you need the single drop letter followed by
% normal text (unknown if ever used by the IEEE):
% \IEEEPARstart{A}{}demo file is ....
% 
% Some journals put the first two words in caps:
% \IEEEPARstart{T}{his demo} file is ....
% 
% Here we have the typical use of a "T" for an initial drop letter
% and "HIS" in caps to complete the first word.

\IEEEPARstart{T}{he} computational costs of convolutional neural networks (CNNs) have increased as CNNs get wider and deeper to perform better predictions for a variety of applications. For deep learning to have revolutionary impact on real-world applications, their computational costs must meet the timing, energy, monetary, and other design constraints of the deployed services. Many approaches have been studied to reduce the computational costs at all levels of software and hardware, from advances in network architectures \cite{szegedy2015going,chollet2017xception} down to electronics where even memory devices have been extensively researched \cite{sun2018fully, shim2016low}.

Although training requires more computations when compared to inference, it is still important to reduce the cost of inference as much as possible because it is the inference that is usually subject to more strict real-world design constraints.
Many hardware-based approaches have shown significant improvements for the computational costs of CNN inferences, but there are two limitations commonly found in these works.
Some techniques are computationally expensive in order to optimize their methods for each network model, or to retrain networks to compensate for the performance degradation from their methods \cite{kung2015power, zhang2015approxann}. 
Also, many techniques such as \cite{rastegari2016xnor} are only effective for small networks and cannot scale to deeper CNNs as they report much worse performance results when tested for deeper networks.
They leverage the fact that a small number of bits are sufficient for small CNNs, but more complex networks require more bits to properly represent the amount of information \cite{lai2017deep}.

One promising hardware-based approach is the application of approximate multiplication to CNN inference \cite{kim2018efficient}. It involves designing and applying multiplication circuits that have reduced hardware costs but produce results that are not exact. Unlike aggressive quantization that trades off numeric precision, the multipliers trade off arithmetic accuracy that is less dependent on the network models, making them better suited for deeper CNNs. The approach does not involve any optimization to a target network model or require additional processing of the network models, allowing easy adaptation into the ASIC and FPGA accelerators.

While optimizing CNN inference through approximate multiplication was demonstrated in several previous studies, there was limited understanding of why it worked well for CNNs. The promising results led to the general observation that CNNs were resilient against small arithmetic errors, but none of them identified the complete reason behind that resilience. Specifically, it was unclear how the CNN layers preserved their functionalities when all their multiplications have a certain amount of error. The lack of understanding made it challenging to identify the suitable approximate multiplier for each network model, leading to expensive search-based methodologies in some studies \cite{mrazek2016design}.

This paper investigates how the errors from approximate multiplication affect deep CNN inference. The work is motivated by hardware circuits but it focuses on the implications from the Deep Learning perspective. 

The contributions are summarized as follows:
\begin{itemize}
    \item Explaining how convolution and fully-connected (FC) layers maintain their intended functionalities despite approximate multiplications.
    \item Demonstrating how batch normalization can prevent the buildup of error in deeper layers when its parameters are properly adjusted. 
    \item Discussing how these findings also explain why bfloat16 multiplication performs well on CNNs despite the reduction of precision.
    \item Performing experiments to show that deep CNNs with approximate multiplication perform reasonably well.
    \item Discussing the potential cost benefits of the methodology by briefly comparing the hardware costs against those of bfloat16 arithmetic.
\end{itemize}

\section{Preliminaries}
\label{sec:prelim}

The convolution layers in CNNs consist of a large number of multiply-accumulate (MAC) operations and they take up the majority of computations for CNN inferences \cite{qiu2016going}.
The MAC operations are ultimately performed in the hardware circuits, and it is important to minimize the cost of these circuits to perform more computations with the same amount of resources.
For MAC operations, multiplications are more complex than additions and consume most resources.
The proposed methodology consists of minimizing the cost of multiplication by replacing the conventional multipliers with approximate multipliers.

Approximate multipliers are significantly cheaper compared to the exact multipliers but they introduce errors in the results.
There are many different types of approximate multipliers with various costs and error characteristics.
Some designs use the electronic properties \cite{chippa2010scalable} and some approximate by intentionally flipping bits in the logic \cite{du2014leveraging}, while others use algorithms to approximate multiplication \cite{sarwar2016multiplier}.

\begin{figure}[t]
\vskip 0.2in
\begin{center}
\centerline{\includegraphics[width=\columnwidth]{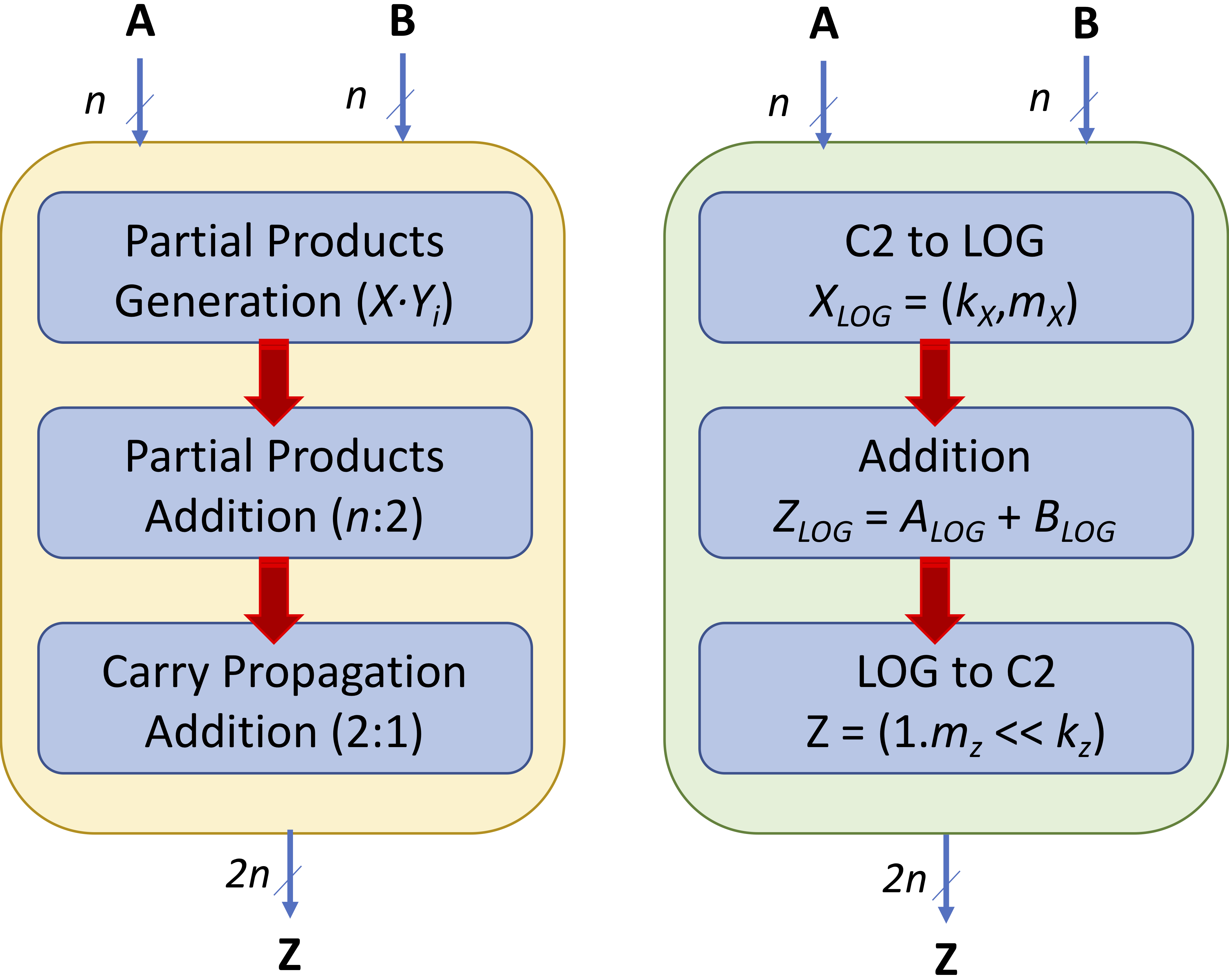}}
\caption{Difference between (a) the conventional fixed-point multiplication and (b) the approximate log multiplication. $k$ stands for characteristic and $m$ stands for mantissa of logarithm.}
\label{fig:logmult}
\end{center}
\vskip -0.2in
\end{figure}

This paper studies the effects of approximate multiplication with the approximate log multiplier presented in \cite{kim2018efficient} as well as a few other promising designs.
The approximate log multiplication is based on the Mitchell's Algorithm \cite{mitchell1962computer} that performs multiplications in the log domain.
Fig. \ref{fig:logmult} shows the difference between the conventional fixed-point multiplier and the log multiplier.
An important benefit of the algorithm-based approximation is the consistent error characteristics which allow for consistent observation of the effects across various CNN instances.
The other types of approximation have more inconsistent errors that make them ill-suited for the study.
For example, approximate multipliers based on electronic properties depend not only on the operands but also on Process, Voltage, and Temperature (PVT) variations, making it difficult to get consistent observations.
The findings of this study are not limited to log multiplication, however, and may help explain the viability of other approaches when they meet the conditions discussed in Section \ref{subsec:minimized}.

\begin{figure}[t]
\captionsetup{justification=centering}
\begin{subfigure}[t]{0.97\linewidth}
\centering
\includegraphics[height=2.3in]{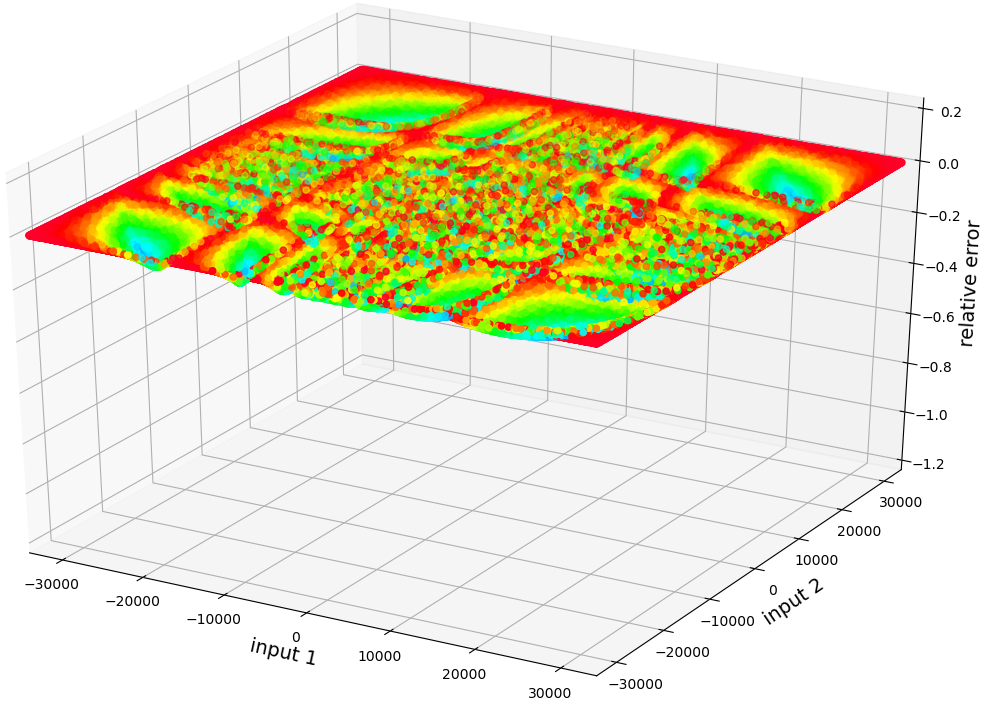}
\caption{Error pattern of the original Mitchell multiplier with exact sign handling, given two signed inputs.}
\label{fig:mitch_pattern}
\end{subfigure}
~
\qquad
\begin{subfigure}[t]{0.97\linewidth}
\centering
\includegraphics[height=2.3in]{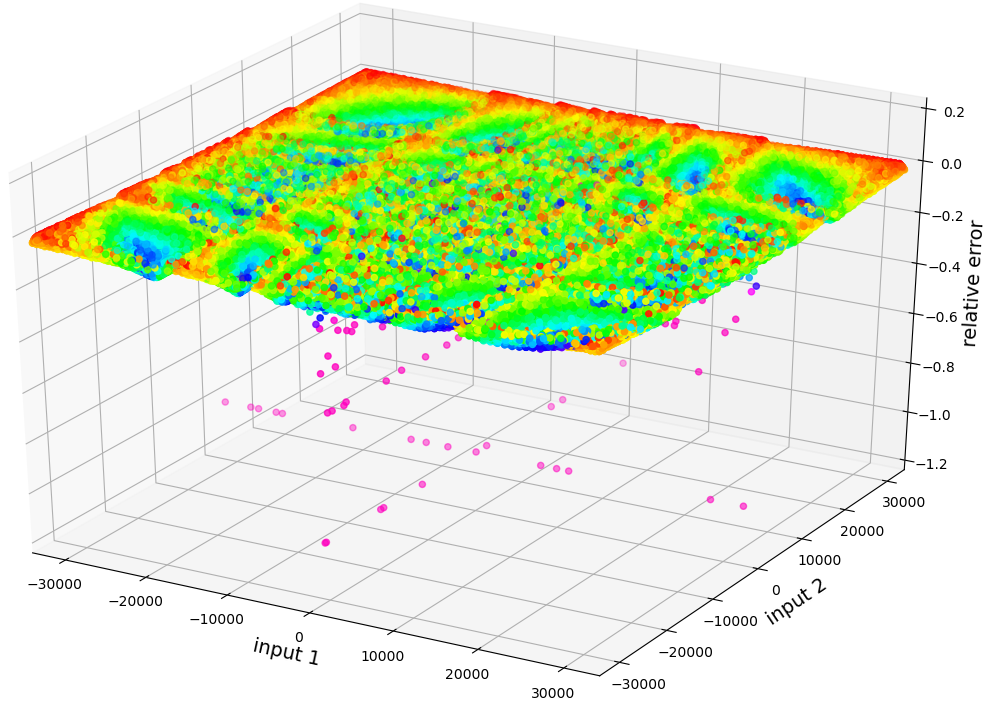}
\caption{Error pattern of Mitch-$w$6 with C1 approximated sign handling, given two signed inputs.}
\label{fig:mitchk_c1_pattern}
\end{subfigure}
~
\qquad
\begin{subfigure}[t]{0.97\linewidth}
\centering
\includegraphics[height=1.9in]{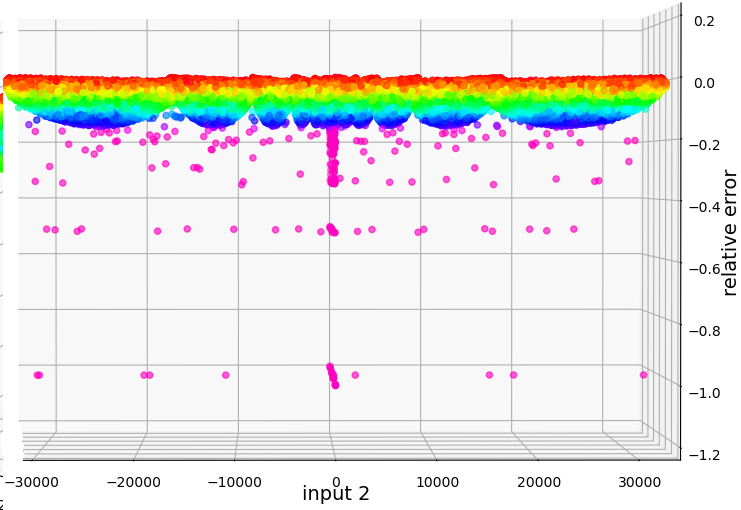}
\caption{Error pattern of Mitch-$w$6, viewed from the side.}
\label{fig:mitchk_c1_pattern_side}
\end{subfigure}
~
\caption{Error patterns of approximate log multipliers.}
\label{fig:error_patterns}
\end{figure}

\begin{figure*}[ht]
\begin{subfigure}[t]{0.24\textwidth}
\centering
\includegraphics[height=3.2in]{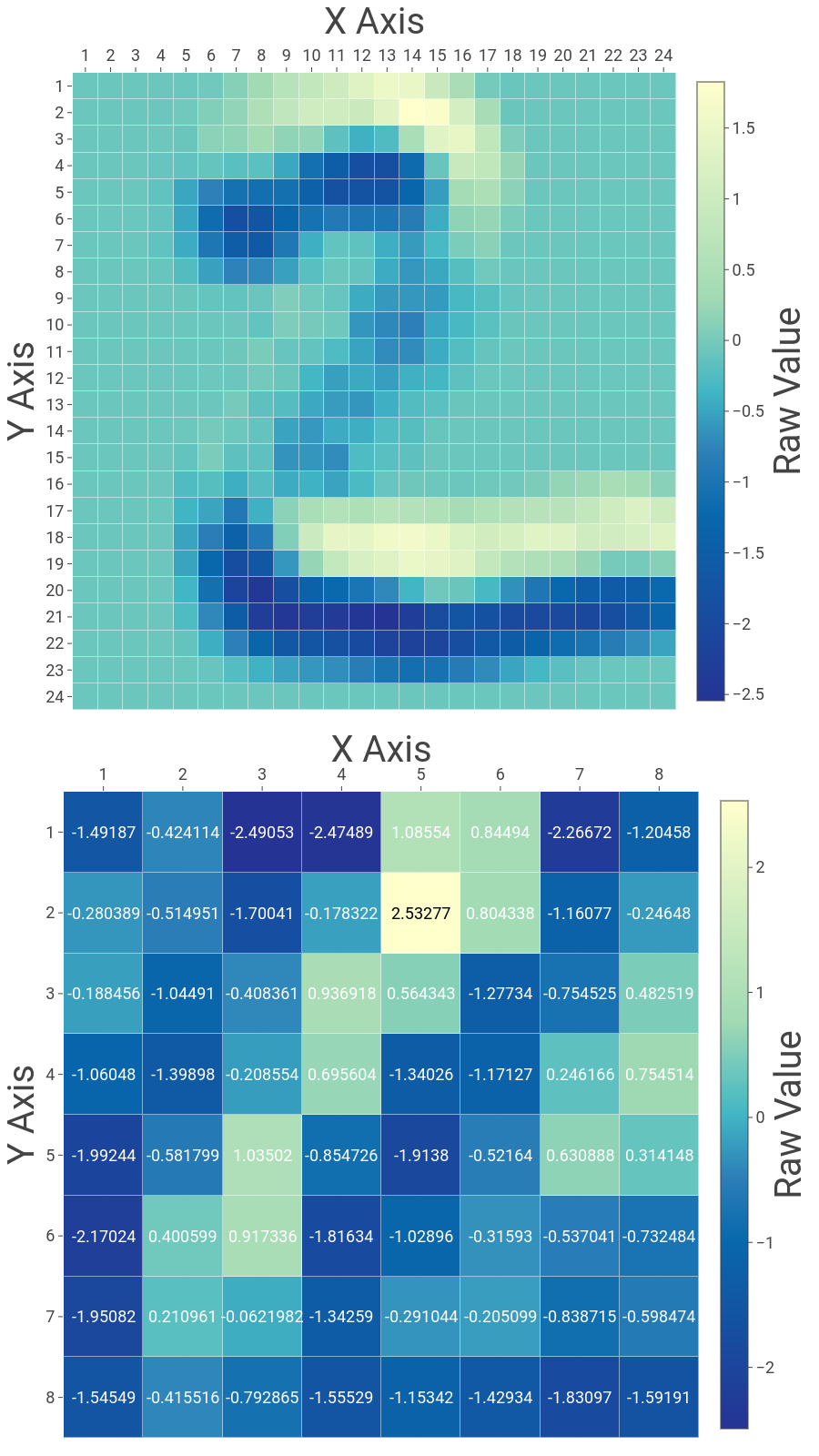}
\caption{Convolution by Log Mult.}
\label{fig:insidemnist_mitchconv}
\end{subfigure}
\qquad
\begin{subfigure}[t]{0.24\textwidth}
\centering
\includegraphics[height=3.2in]{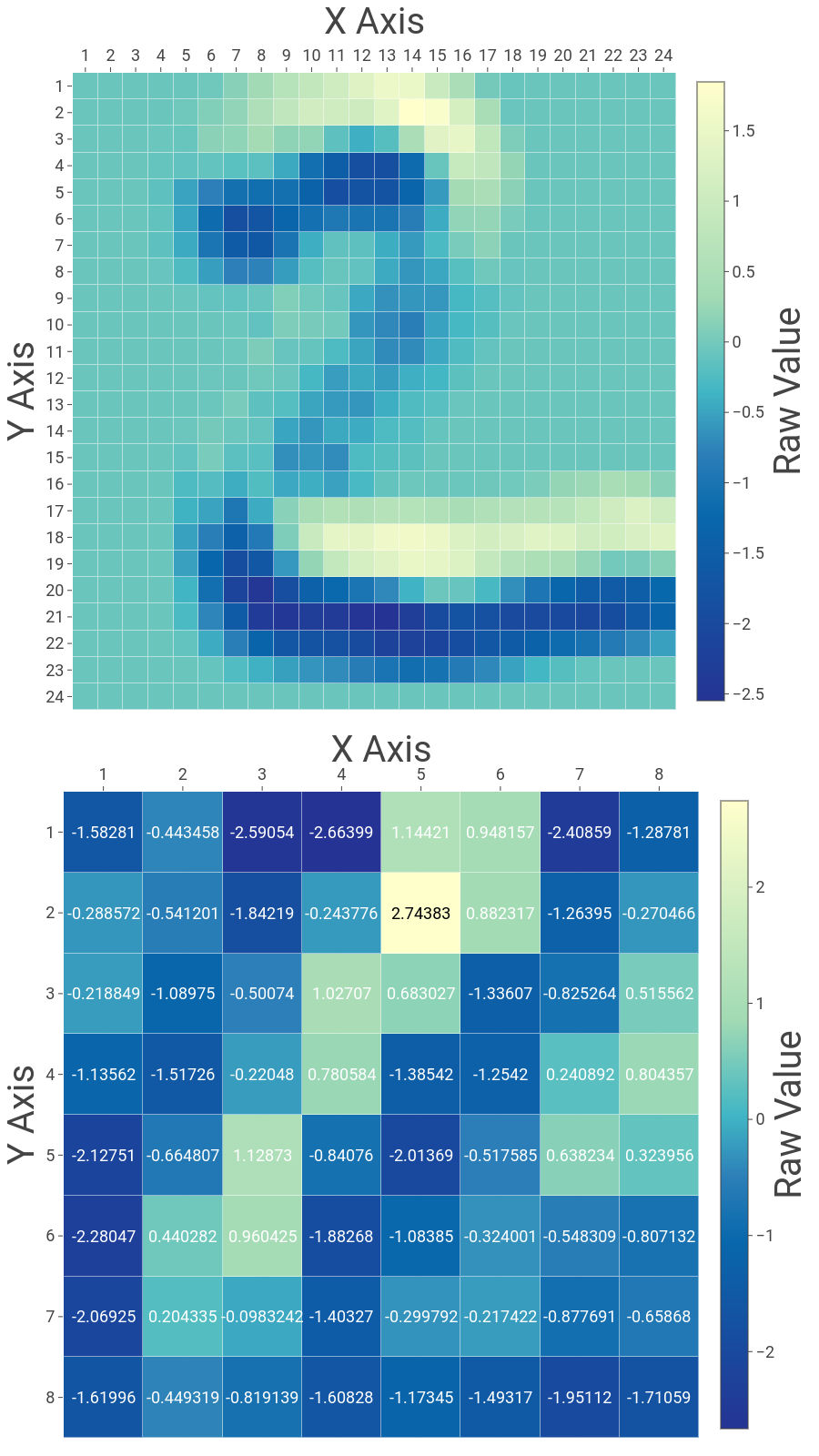}
\caption{Convolution by Float Mult.}
\label{fig:insidemnist_convfloat}
\end{subfigure}
\qquad
\hspace{-5mm}
\begin{subfigure}[t]{0.48\textwidth}
\centering
\includegraphics[height=3.2in]{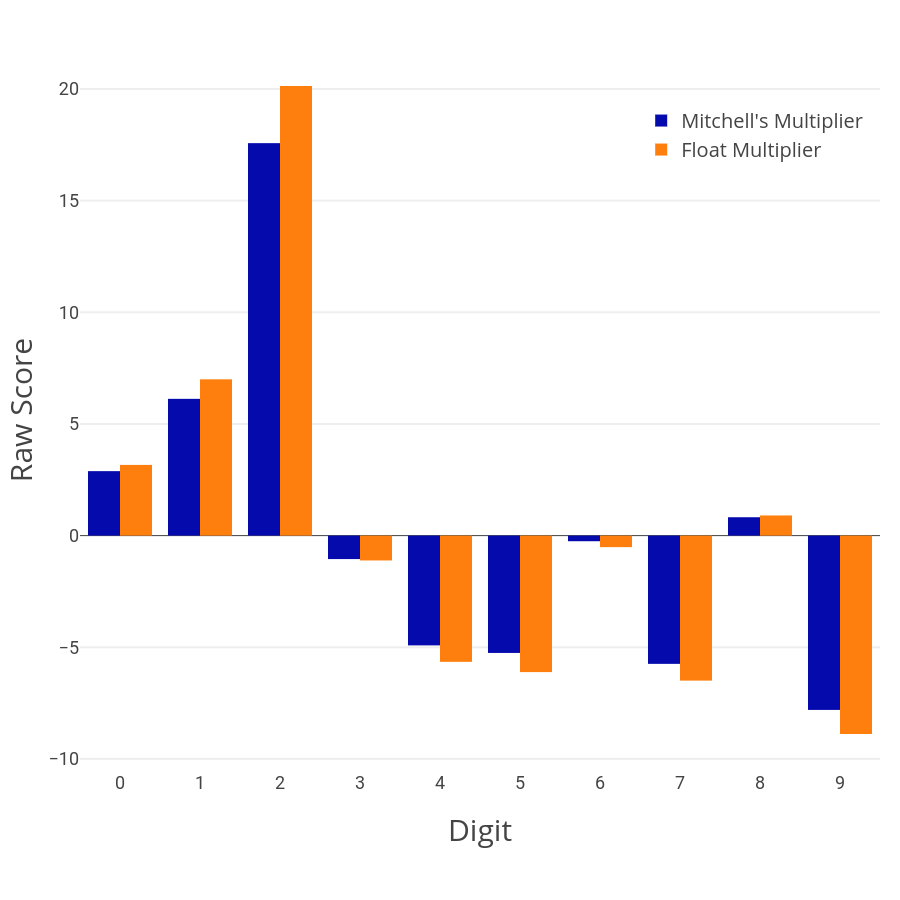}
\caption{The final scores}
\label{fig:insidemnist_finalscore}
\end{subfigure}
\caption{Convolution outputs and the final raw scores of a sample inference from LeNet \cite{kim2018efficient}.}
\label{fig:insidemnist}
\end{figure*}

The errors from approximate log multiplication are deterministic and depend on the two input operands, similarly to the other algorithmic approximation methods. Fig. \ref{fig:error_patterns} shows the error patterns of the original Mitchell log multiplier \cite{mitchell1962computer} and Mitch-$w$6 \cite{kim2018efficient} with a million random input pairs. The relative error is defined as Equation \ref{eq:deferr} where $|Z|$ is the magnitude of the exact product and $|Z'|$ is the magnitude of the approximate product. 
\begin{align} \label{eq:deferr}
\begin{split}
error_{relative} = \frac{|Z'|- |Z|}{|Z|}.
\end{split}
\end{align}
Approximate log multiplication requires separate sign handling and does not affect the signs of the products \cite{kim2018efficient}.
Compared to the original Mitchell log multiplier, Mitch-$w$6 has a small frequency of high relative errors caused by the 1's complement (C1) sign handling, but they are acceptable as CNNs consist of MAC operations \cite{kim2018efficient}.
It should be noted that the approximate log multipliers have reasonably even distributions of errors across the input ranges, but can only have negative errors that cause the products to have less magnitudes compared to the exact products.
The mean error of an approximate multiplier is measured by repeating many multiplications with random inputs, and the Mitchell multiplier has the biased mean error of -3.9\% at 32 bits while Mitch-$w$6 has -5.9\%.

Besides the convolution layers, the FC layers also have MAC operations but they have fewer computations compared to convolution \cite{qiu2016going}.
Our methodology still applies to approximate multiplication of FC layers to be consistent with networks that use 1x1 convolution for classifiers.
The effect of approximating FC layers is minimal because of the reasons discussed in Section \ref{sec:errorconv}. 
On the other hand, the operations in batch normalization are not approximated because they can be absorbed into neighboring layers during inferences \cite{lin2016fixed}.

It is important to understand the difference between the method of quantization and the approximate multiplication.
Quantization is the process of converting floating-point values in the CNN models to fixed-point for more cost-efficient inferences in the hardware \cite{lin2016fixed}. 
The goal of quantization is to find the minimum number of fixed-point bits that can sufficiently represent the distribution of values. In fact, there are some approximations with small numbers of fixed-point bits that cannot match the range and precision of the floating-point format. The error from this approximation depends on the network models as each has different distributions of values \cite{judd2016stripes,lai2017deep}. The network dependency is the reason why more complex networks require a higher number of bits and the benefits of aggressive quantization diminish. While many studies have successfully demonstrated the effectiveness of quantization, they usually report significant degradation of CNN prediction accuracies when using only 8 bits on deep CNNs \cite{jacob2018quantization}.

Approximate multiplication is less dependent on the networks because its source of error is from the approximation methods, not any lack of range and precision. 
Given proper quantization, the approximate multiplication further minimizes the cost of multipliers for the given number of bits.
Approximate multiplication is an orthogonal approach to quantization as approximate multipliers may be designed for any number of bits, and it complements quantization to maximize the computational efficiency of CNN inferences.

\section{Accumulated Error in Convolution}
\label{sec:errorconv}

This section explains how the convolution and FC layers achieve their intended functionalities despite the errors from approximate multiplication.

\subsection{Understanding Convolution and FC Layers}

Explaining the effects of approximate multiplication must begin with understanding how the convolution and FC layers achieve their intended functionalities.
Fig. \ref{fig:insidemnist} is taken from \cite{kim2018efficient} and shown here to visualize the outputs of convolution and FC.
The CNN convolution layers achieve abstract feature detection by performing convolution between their input channels and kernels.
They produce feature maps, as shown in Fig. \ref{fig:insidemnist_mitchconv} and \ref{fig:insidemnist_convfloat}, where the locations that match the kernel are represented by high output values relative to other locations.
Unlike a sigmoid or step activation, the widely used ReLU activation function simply forces the negative output values to zero and does not have absolute thresholds with which the abstract features are identified.
That means the abstract features are not identified by their absolute values but by the relatively higher values within each feature map, and this claim is also supported by the fact that convolution is often followed by a pooling layer.
Similarly, when the FC layers classify an image based on the abstract features, the probabilities of classes are decided by the relative strengths and order among all FC outputs.
CNNs simply select the best score as the most probable prediction instead of setting a threshold with which a prediction is made.

Because the features are represented with relative values as opposed to absolute values, it is much more important to minimize the variance of error between the convolution outputs than minimizing the absolute mean of errors when applying approximate multiplication to convolution \cite{kim2018efficient}. 
In other words, it is acceptable to have a certain amount of error in multiplications as long as the errors affect all outputs of convolution as equally as possible.
The FC layers behave in the same way so that it is important to minimize the variance of error between the nodes.
Fig. \ref{fig:insidemnist} demonstrates this principle and shows that the Mitchell log multiplier can produce a correct inference because all outputs are affected at the same time.
Fig. \ref{fig:insidemnist} also shows that the variances of accumulated errors in the convolution and FC layers are very small when the approximate log multiplier is applied, and the convolutions are still able to locate the abstract features albeit with smaller magnitudes. 
The previous work \cite{kim2018efficient}, however, did not identify the reason why the variance of accumulated error was minimized when approximate multiplication was applied.

\subsection{Minimized Variance of Error}
\label{subsec:minimized}

This paper provides the analytical explanation for why the variance of accumulated error was minimized in the convolution and FC layers. These layers consist of large numbers of multiplications and accumulations that converge the accumulated errors to a mean value. The variance of the accumulated error is minimized and all outputs of the layers are equally affected because of this convergence, and then maintaining the relative magnitudes between the outputs preserves the functionality of abstract feature detection.

Equation \ref{eq:multiconv} shows the multi-channel convolution where feature $s$ at ($i$,$j$) is the accumulation of products between kernel $w$ and input $x$ across the kernel dimensions ($m$,$n$) and the input channels ($l$).
\begin{equation}
\label{eq:multiconv}
s_{i,j} = \sum_{l}\sum_{m}\sum_{n} w_{l,m,n} \cdot x_{l,i-m,j-n}~.
\end{equation}

\begin{figure}[t]
\vskip 0.2in
\begin{center}
\centerline{\includegraphics[height=1.8in]{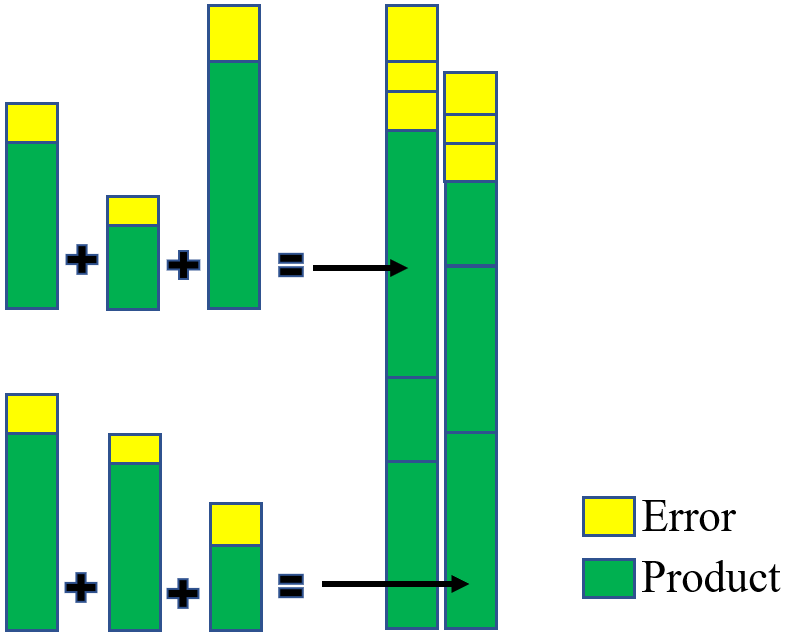}}
\caption{Accumulation of many products with varying amount of error converges the combined errors to a mean value.}
\label{fig:converge}
\end{center}
\vskip -0.2in
\end{figure}

The distributions of weights and inputs are different for each CNN model and layer \cite{qiu2016going,lai2017deep,judd2016stripes}.
The input operands to multiplication, weights and input pixels, are numerous and practically unpredictable with pseudo-randomness, which in turn makes the error from approximate multiplication pseudo-random.
The approximate log multipliers have evenly distributed error patterns across the input ranges, as shown in Fig. \ref{fig:error_patterns}, and therefore the expected value of the error is close to the mean error of the approximate multiplier regardless of the different ranges of inputs from CNNs.
When each convolution output accumulates many products from approximate multiplication, the accumulated error statistically converges closer to the expected value, which is the mean error of the approximate multiplier.
This convergence reduces the variance of the accumulated error between the outputs and the values scale by roughly the same amount, minimizing the effect of varying error on feature detection.
Fig. \ref{fig:converge} shows the abstraction of this mechanism and Fig. \ref{fig:insidemnist} shows an example. 
Equation \ref{eq:converr} describes the feature $s'_{i,j}$ when multiplications are associated with the mean error of $e$.
\begin{gather}
\label{eq:converr}
s'_{i,j} = \sum_{l}\sum_{m}\sum_{n} w_{l,m,n} \cdot x_{l,i-m,j-n} \cdot (1 + e)~, \\
\label{eq:converrfinal}
s'_{i,j} = (1+e) \cdot s_{i,j}~.
\end{gather}
Therefore, the features are simply scaled by the mean error of the approximate multiplication when a large number of products are accumulated.

The above observations hold only for the approximate multiplications with the symmetric errors between positive and negative products so that Equations \ref{eq:converr} and \ref{eq:converrfinal} hold. 
The approximate multipliers studied in this paper satisfy this condition because all of them handle the signs separately from magnitudes.

Although we primarily used the Mitch-$w$ multiplier to develop this hypothesis, the hypothesis does not depend on the inner workings of the log multiplier but only on the output error characteristics. Therefore, the theory can be similarly applied to any approximate multiplier that meets the assumptions made in this section, namely the evenly distributed error and the symmetric errors between positive and negative products. Having only negative errors like Mitch-$w$ is not a requirement. It should be noted that the assumption of an evenly distributed error is used to accommodate different ranges of inputs, and may be relaxed when an approximate multiplier can produce a consistent expected value of error for particular input distributions. In this paper, we also used DRUM6 \cite{hashemi2015drum} and the truncated iterative log multiplier \cite{kim2019cost} for the experiments in Section \ref{sec:experiments} to show that the hypothesis may be applied to other approximate multipliers.

%Although we primarily used the log multipliers to present this theory, it is not limited to log multiplication and any approximate multiplier with the evenly distributed error and the symmetric sign behavior are eligible.
%Notably, DRUM6 \cite{hashemi2015drum} was used along with the log multipliers to provide the experimental results in Section \ref{sec:experiments}.

\subsection{Impact on Convolution and FC}
\label{subsec:impact_on_conv}

\begin{figure}[t]
\vskip 0.2in
\begin{center}
\centerline{\includegraphics[height=2.4in]{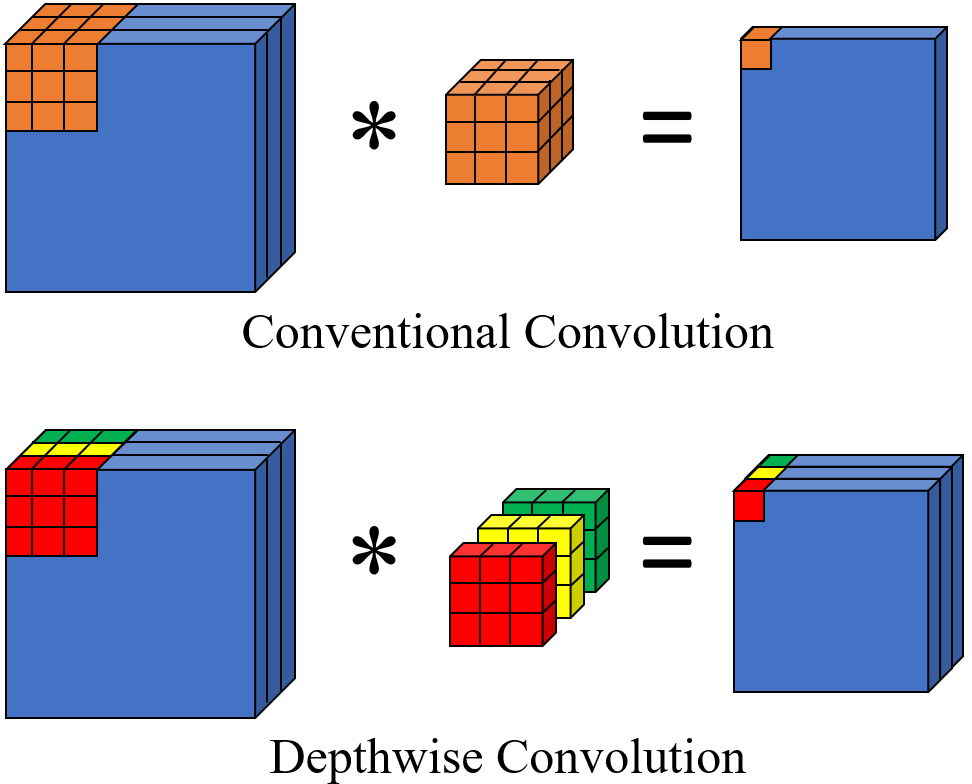}}
\caption{Depthwise Convolution has a reduced number of accumulations and convergence of error per output.}
\label{fig:diffconv}
\end{center}
\vskip -0.2in
\end{figure}

The number of accumulations in convolution is finite so the convergence does not completely nullify the variance of accumulated error.
The small amount of error variance from approximate multiplication is acceptable, however, because CNNs are designed to be general and robust against small variations by nature.
The techniques of regularization, such as pooling and dropout, intentionally lose some information to suppress overfitting and increase the generality of CNN predictions. Some studies have observed that small arithmetic errors have similarly positive effects \cite{ansari2019improving,kim2018efficient,wang2019bfloat16}.
For example, an eye needs to be recognized as an eye even when it is a little different from the training samples. CNNs are designed to overlook such small differences, and some computational inaccuracies are not only tolerable but often beneficial in providing such generality.

Deep CNNs typically start with smaller numbers of convolution channels to obtain general features, and the number of channels increases in the deeper layers where features become more specific.
Approximate multiplication on such CNNs exhibits the desired trend of having smaller effects in the wide and deep layers as required.
The larger variance of accumulated error in the shallow layers is tolerable because the feature detection needs to account for the small variations in the input images.
In fact, some previous works, such as \cite{sarwar2016multiplier, kim2017power}, had claimed that earlier layers can be approximated more in neural networks.

This hypothesis implies the importance of exact additions in CNNs because the multiplication errors will not converge properly with inexact accumulations.
This agrees with the work in \cite{du2014leveraging} where approximating the additions had a larger impact on the CNN accuracies.
As multipliers in fixed-point arithmetic are much more expensive than adders, approximating only the multipliers gains the most benefit with minimal degradation in CNN inferences.

Approximate multiplication also benefits from the fact that the convolution outputs receive inputs from the same set of input channels.
For each convolution output, there are two types of accumulations. 
One type occurs within each input channel across the kernel dimensions while the other occurs across the input channels to produce the final output. 
The intra-channel accumulation combines the products from the same input channel and kernel, and therefore each channel has a specific range of values within which features are located. 
The inter-channel accumulation may have more varying ranges of products because each input channel has its own kernel and input values.
Different input ranges may trigger different error characteristics on the approximate multiplier, but every convolution output accumulates from all input channels so that it does not affect the variance of accumulated error between the outputs.
An implication of this observation is that approximate multiplication does not work as well when every output does not accumulate from the same set of data, as in the cases of grouped convolution and branches in CNN architectures.

The FC layers are also resilient against the effects of approximate multiplication as the same factors help converge errors in the outputs.
There is usually a large number of accumulations per each output and all outputs share the same set of inputs.
Thus, CNN accuracies show minimal differences when the FC layers have approximate multiplications as demonstrated in Section \ref{sec:experiments}.

\subsection{Grouped and Depthwise Convolutions}
\label{subsec:grouped}

\begin{figure*}[ht]
\vskip 0.2in
\begin{center}
\centerline{\includegraphics[height=1.6in]{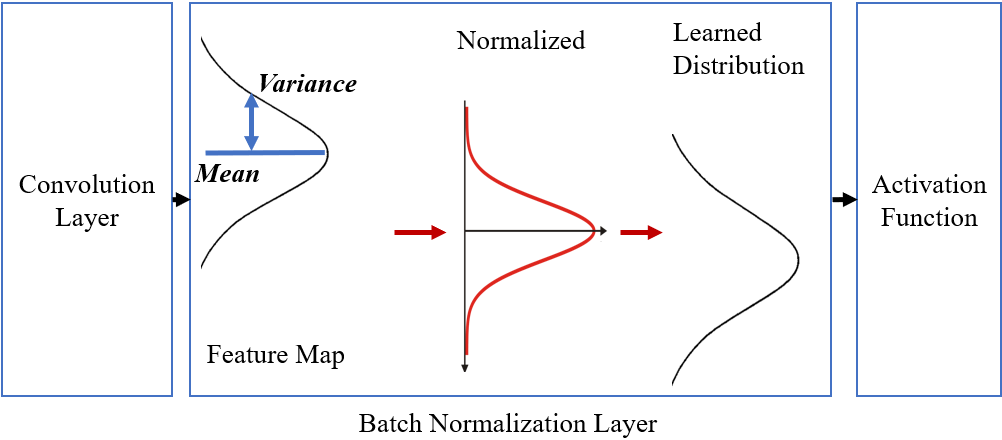}}
\caption{Abstract overview of batch normalization.}
\label{fig:batchnorm}
\end{center}
\vskip -0.2in
\end{figure*}

The benefits of approximate multiplication with conventional convolution are best understood and verified by comparing against grouped and depthwise separable convolution. 
Depthwise separable convolution consists of depthwise convolution followed by pointwise convolution \cite{chollet2017xception}.
Depthwise convolution is a special case of grouped convolution that eliminates the accumulation across input channels, and the reduced number of accumulations leads to an increase in the variance of accumulated error in the outputs.
Fig. \ref{fig:diffconv} shows the comparison of the accumulation pattern between conventional convolution and depthwise convolution.
Also, each output channel receives inputs from only one input channel and the difference of error between output channels is subject to another approximate multiplication and variance of error before the inter-channel accumulations occur in the following pointwise convolution.
More accurate approximate multipliers are required for CNNs that use depthwise separable convolution because errors from approximate multiplication do not converge well.
A sufficiently accurate approximate multiplier can still perform reasonably well, as demonstrated in Section \ref{sec:experiments}.

%One may argue that the incompatibility with depthwise separable convolution makes approximate multiplication impractical, because depthwise separable convolution is a technique that significantly reduces the amount of computations.
%This incompatibility does not diminish the significance of this study for several reasons.
%Firstly, depthwise separable convolution does not supplant the conventional convolution.
%Depthwise separable convolution has fewer learnable parameters and CNNs that use it tend to have less prediction capabilities.
%Also, approximate multiplication reduces the computing resources in a very different way compared to depthwise separable convolution.
%While depthwise separable convolution reduces the number of MAC operations, approximate multiplication reduces the cost of each MAC operations, and each will behave differently depending on various design choices.
%Lastly, approximate multiplication and depthwise separable convolution are not strictly incompatible, because 

Another technique that reduces the number of accumulations is 1x1 convolution, but it is found to be compatible with approximate multipliers.
1x1 convolution does not have any intra-channel accumulation but accumulates the products across input channels.
Because deep CNNs require large numbers of channels appropriate for their deep structures, inputs to 1x1 convolutions usually consist of many input channels and therefore provide enough accumulations for the error convergence.
Each output of 1x1 convolution also receives inputs from all input channels, which provides more consistent accumulation of error between the outputs.

\section{Effect of Batch Normalization}
\label{sec:batch}

The approximate log multiplication with Mitchell's Algorithm generates negative error in the results, meaning that the product has less magnitude compared to the exact multiplication \cite{mitchell1962computer}.
It is evident from Equation \ref{eq:converrfinal} that the features have less magnitudes with the log multiplication in each convolution layer. 
There are many convolution layers that repeatedly cause the reduction, and the previous work had reported that this became a problem for deeper layers \cite{kim2018efficient}.
Its adverse effect on the network performance was observable in AlexNet with only 8 layers of convolution and FC, and it was unclear how the mean error accumulation would behave in much deeper networks.
Having tens or hundreds of convolution layers significantly reduces the magnitudes of the features so that the deeper layers receive input distributions that are difficult to distinguish. 
On the other hand, if an approximate multiplier has a positively biased mean error, it is possible to amplify the values beyond the range set by quantization, resulting in the arithmetic overflow.
These adverse effects are under the best-case scenario of ReLU activation, and the other types such as a sigmoid function may suffer additional errors in activations.
The ReLU function simply forces the negative values to zero and does not change the magnitudes of positive inputs, but the same is not true for other activation functions where the magnitudes of positive inputs cause changes in activations.

Batch normalization \cite{ioffe2015batch}, the popular technique used in most deep CNNs, can alleviate this problem and help approximate multiplication go deeper into the networks.
A critical function of batch normalization is to redistribute the output feature maps to have more consistent input distributions for deeper layers.
While the training process necessitates this function, the inferences on the resulting models still need to go through the normalization with the stored global parameters of expected distributions.
These global parameters can be appropriately adjusted to account for the changes in the distributions due to approximate multiplication, and this can prevent the accumulation of mean error across the layers.

The abstract overview of batch normalization is shown in Fig. \ref{fig:batchnorm}.
During training, each batch normalization layer calculates and stores the mean and variance values of the input distributions.
These mean and variance values are used to normalize the input distributions to generate the normalized distributions with the mean value of zero and the variance of one.
Then, batch normalization uses learnable parameters to scale and shift the normalized distribution to restore the representation power of the network \cite{ioffe2015batch}.
In essense, batch normalization redistributes the feature maps before or after the activation function so that the next layer may receive consistent distributions of inputs.
All these parameters are learned during training and stored as numerical values in CNN models, and they can be easily modified if necessary.
CNN inferences use these stored parameters to perform normalization assuming they represent the same input distributions during inferences.

The mean and variance parameters are a source of error for approximate multiplication without proper adjustments because the distribution of convolution outputs changes as the result of approximate multiplication. 
Equations \ref{eq:meanfeat3} and \ref{eq:varifeat3} show the mean ($\mu'$) and variance ($(\sigma')^2$) of the convolution output distribution, when the features $s'_{i,j}$ have the mean error $e$ from Equation \ref{eq:converrfinal}.
\begin{gather}
\label{eq:meanfeat1}
\mu' = 1/ m \sum_{i,j} s'_{i,j}~,\\
\label{eq:meanfeat2}
\mu' = 1/m \sum_{i,j} (1+e) \cdot s_{i,j}~,\\
\label{eq:meanfeat3}
\mu' = (1 + e) \mu~.\\
\label{eq:varifeat1}
(\sigma')^2 = 1/m \sum_{i,j} (s'_{i,j} - \mu')^2~,\\
\label{eq:varifeat2}
(\sigma')^2 = 1/m \sum_{i,j} (1 + e)^2 (s_{i,j} - \mu)^2~,\\
\label{eq:varifeat3}
(\sigma')^2 = (1 + e)^2 \cdot \sigma^2~.
\end{gather}
Therefore, the stored mean values for batch normalization must be scaled by $(1 + e)$, while the variance values are scaled by $(1+e)^2$.
With the adjusted parameters, the batch normalization layers correctly normalize the convolution outputs and scale them back to the desired distributions.
In the process, the mean and variance of the outputs match those of exact multiplication and the effect of mean error accumulation disappears.
Failing to adjust these parameters results in incorrect redistribution of feature maps, and worse CNN accuracies.
The proposal only requires the scaling of the stored parameters and significantly improves the performance of approximate multipliers on deep neural networks.
It does not introduce any new operations and does not prevent the ability of batch normalization to fold into neighboring layers.

Designing an approximate multiplier with an unbiased mean error near zero is another effective solution, but it is much harder to make changes to hardware designs.
The unbiased designs usually have a small amount of mean error because it is difficult to create a perfectly unbiased design, and the problem is only deferred to deeper networks.
Also, depending on the approximation method, it may take additional hardware resources to make a design unbiased.
The networks that do not use batch normalization have no choice but to use the unbiased multipliers, but otherwise the proposed adjustment is simpler, less costly, and more flexible to accommodate different approximation methods with biased mean errors.

\section{Arithmetic Reason for Bfloat16 Success}
\label{sec:bfloat16}
The discoveries in Sections \ref{sec:errorconv} and \ref{sec:batch} are not limited to the error of approximate multiplication but apply to all sources of arithmetic error.
They also provide deeper understanding of why bfloat16 \cite{wang2019bfloat16} has been widely successful at accelerating CNNs despite its reduced precision.
The bfloat16 format is an approximation of the FP32 floating-point format that simply truncates the 16 least significant bits from the 23 fractional bits.
By truncating the less significant fractional bits, converting an FP32 value to bfloat16 generates a small negative error from 0\% to -0.78\% relative to the original FP32 value. The factors discussed in Section \ref{sec:errorconv} also minimize the adverse effects of this varying error and they explain why using the full FP32 accumulator after bfloat16 multiplication produces the best results \cite{henry2019leveraging}, in agreement with the observation that the accumulations need to be exact.
The accumulation of mean error discussed in Section \ref{sec:batch} should also be present, but the mean error of bfloat16 is too small to cause any problems for the studied CNNs.

The successful application of bfloat16 to CNNs has been explained by the high-level interpretation that the small amount of error helps the regularization of a CNN model. The interpretation is still valid and also applies to approximate multiplication, and the findings of this paper provide deeper understanding with the arithmetic explanation. They also explain why the bfloat16 format has slightly degraded performances with the networks that use grouped convolution as presented in Section \ref{subsec:expvari}.

\section{Experiments}
\label{sec:experiments}

\subsection{Setup}
\label{subsec:setup}
The experiments are performed in the Caffe framework to evaluate the impact of approximate multipliers on deep CNN models \cite{yangqing2014caffe}.
Caffe has limited features compared to contemporary tools but its lack of encapsulation allows easy modification of underlying matrix multiplication, making it suitable for the study.
The code that performs floating-point matrix multiplication in GPU is replaced by the CUDA C++ functions that emulate the behavior of the target approximate multipliers.
These functions are verified against RTL simulations of the HDL code of the multipliers.

The Mitch-$w$6 multiplier with the C1 sign handling is chosen because the comparison against the other multipliers showed that it was cost-efficient while performing well on AlexNet \cite{kim2018efficient}.
Mitch-$w$ multipliers consume significantly less resources compared to the Mitchell log multiplier.
DRUM6 multiplier \cite{hashemi2015drum} is also added to the experiments because it performed very well on AlexNet while being more costly than Mitch-$w$6 \cite{kim2018efficient}.
The truncated iterative log multiplier in \cite{kim2019cost} has higher accuracy than these multipliers and is tested for networks that have depthwise separable convolution. Unlike Mitch-$w$, DRUM6 and the truncated iterative log multiplier have the unbiased mean errors close to zero.
The FP32 floating-point results are included for comparison, and the bfloat16 results provide additional data points (see Section \ref{sec:bfloat16}).

\begin{table}
	\centering
    \scriptsize
	\caption{Pre-trained CNN models used for the experiments}
 
    \vskip 0.1in   
    \renewcommand\tabcolsep{3pt}
    \renewcommand\arraystretch{1.5}
    
	\begin{tabular}{llcc@{\hspace{7.0pt}}ccc}
		\toprule[0.8pt]
		
		\multicolumn{1}{l}{\textbf{Network}} &
		\multicolumn{1}{c}{\textbf{Model Source}} &
		\multicolumn{1}{c}{\textbf{BatchNorm}} &
		\multicolumn{1}{c}{\textbf{Grouped Conv.}} &
		\\
        
        \cmidrule[0.5pt](l{2pt}r{2pt}){1-4}
		
		\textbf{VGG16} &
		\cite{yangqing2014caffe} &
		 &
		 &
		\\
		
		\textbf{GoogLeNet} &
		\cite{yangqing2014caffe} &
		&
		&
		\\
		
		\textbf{ResNet-50} &
		\cite{he2016deep} &
		$\surd$ &
		&
		\\
		
		\textbf{ResNet-101} &
		\cite{he2016deep} &
		$\surd$ &
		&
		\\
		
		\textbf{ResNet-152} &
		\cite{he2016deep} &
		$\surd$ &
		&
		\\
		
		\textbf{Inception-v4} &
		\cite{liu2019enhancing} &
		$\surd$ &
		&
		\\
		
		\textbf{Inception-ResNet-v2} &
		\cite{silberman2017tf} &
		$\surd$ &
		&
		\\
		
		\textbf{ResNeXt-50-32x4d} &
		\cite{Xie2016} &
		$\surd$ &
		$\surd$ &
		\\
		
		\textbf{Xception} &
		\cite{liu2019enhancing} &
		$\surd$ &
		$\surd$ &
		\\
		
		\textbf{MobileNetV2} &
		\cite{sandler2018mobilenetv2} &
		$\surd$ &
		$\surd$ &
		\\
		
		\bottomrule[0.8pt]
	\end{tabular}%
	\label{tab:list_cnn}%
	\vskip -0.1in
\end{table}%

\begin{figure*}[ht]
\vskip 0.2in
\begin{center}
\centerline{\includegraphics[height=2.2in]{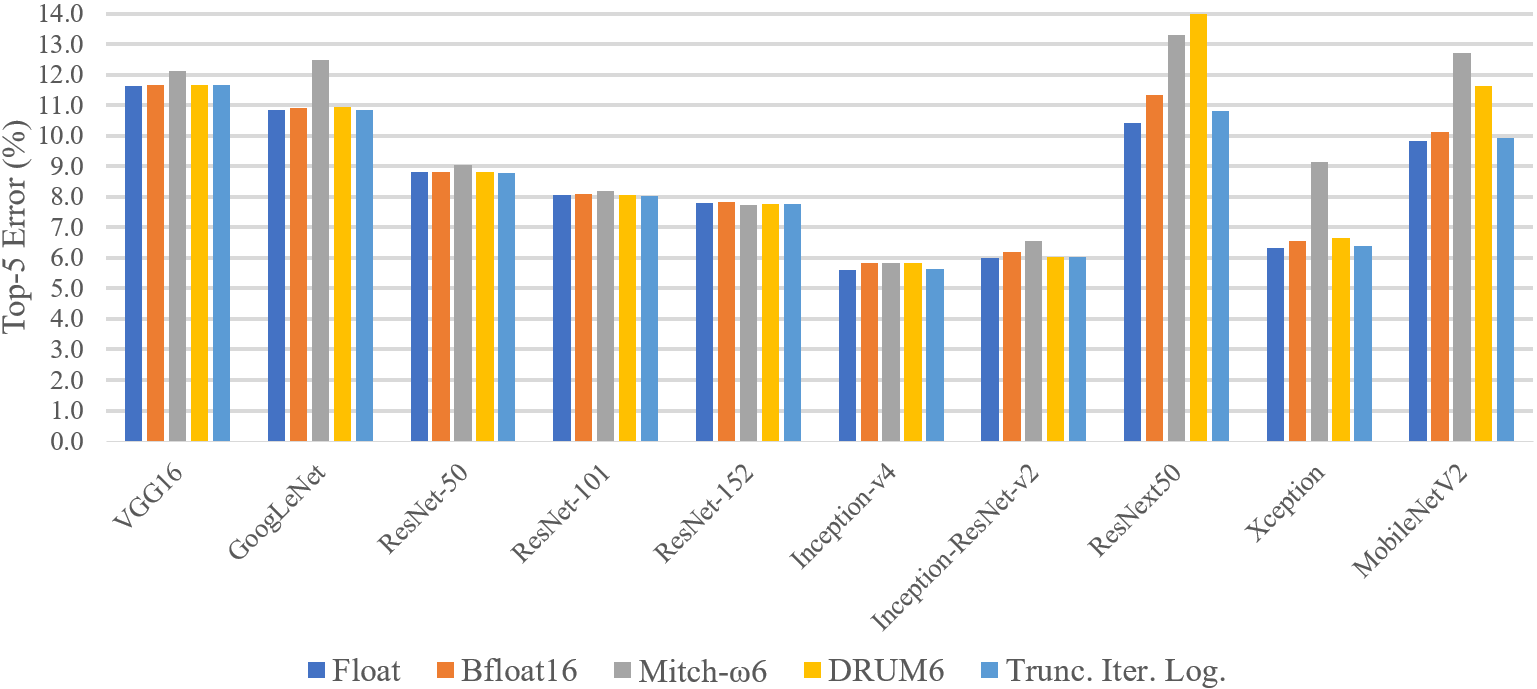}}
\caption{Comparison of Top-5 errors between the FP32 reference and the approximate multipliers.}
\label{fig:top5}
\end{center}
\vskip -0.2in
\end{figure*}

\begin{figure*}[ht]
\vskip 0.2in
\begin{center}
\centerline{\includegraphics[height=2.2in]{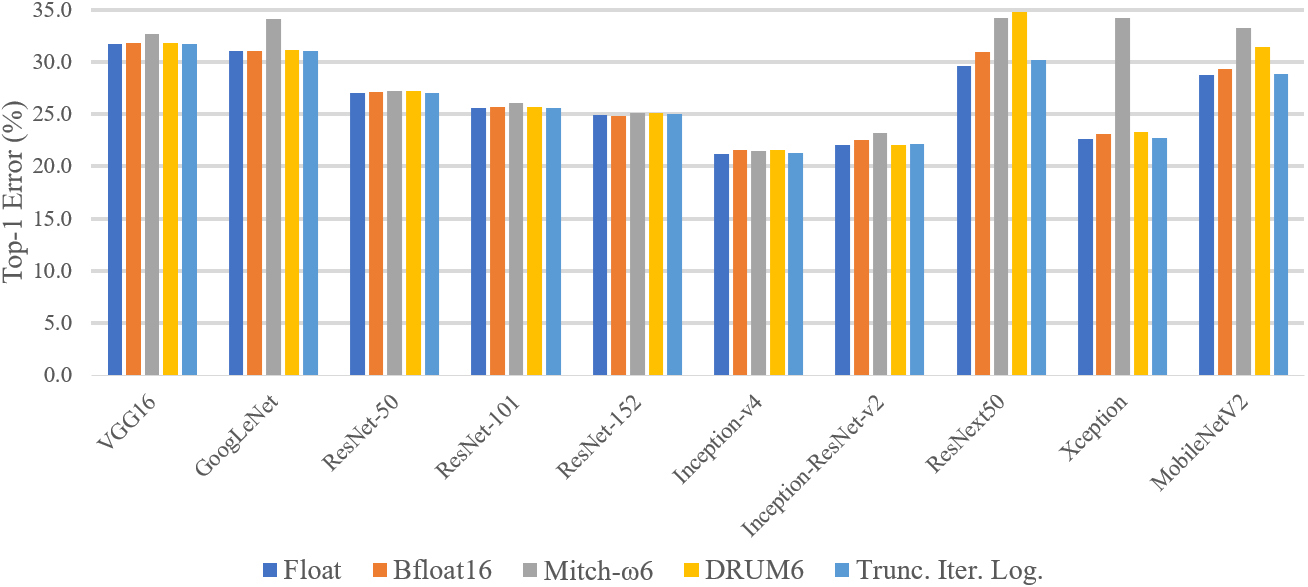}}
\caption{Comparison of Top-1 errors between the FP32 reference and the approximate multipliers.}
\label{fig:top1}
\end{center}
\vskip -0.2in
\end{figure*}

The target application is object classification with the ImageNet ILSVRC2012 validation dataset of 50,000 images. 
Only single crops are used for experiments because the C++ emulation of the approximate multipliers is very time-consuming compared to the multiplication performed in actual hardware, so the presented CNN accuracies may differ from the original literature that use 10-crops. 
Table \ref{tab:list_cnn} shows the list of CNN models used for the experiments, and the networks that use batch normalization and grouped convolutions are marked for comparative discussion. 
The pre-trained CNN models for the experiments are publicly available from online repositories, and the source is indicated with each model.
Any training or retraining of a network model is purposefully avoided to achieve reproducibility and to show that the proposed methodology works with many network models with only minor scaling of batch normalization parameters.

The experiments assume quantization to 32 fixed-point bits without rounding (statically assigned to 16 integer and 16 fractional bits) as it is sufficient for all the tested network models.
As discussed in Section \ref{sec:prelim}, approximate multiplication is an orthogonal approach to quantization and we used generous quantization to minimize the quantization errors and study the effects of approximate multiplication in isolation, in order to clearly evaluate the hypothesis presented in this paper.
This paper focuses on establishing approximate multiplication as a viable approach, and combining various quantization methods with approximate multiplication is beyond the scope of this paper.
%and performing careful quantization to each network model may achieve a lower number of bits.

\subsection{Impact of Approximate Multiplication on CNNs}
\label{subsec:expvari}

Fig. \ref{fig:top5} and \ref{fig:top1} show the Top-5 and Top-1 errors when the approximate multipliers are applied to the CNNs, compared against the FP32 reference values.
For the networks with conventional convolution, the studied approximate multipliers produce predictions that are nearly as accurate as the exact FP32 floating-point as they show Top-5 errors within 0.2\% compared to the reference values, except for Mitch-$w$6 on Inception-ResNet-v2 (0.5\%) and the networks without batch normalization.
On the contrary, the CNNs with grouped convolution suffer degraded accuracies when there are errors in multiplications, from approximate multiplication as well as bfloat16.
The difference of CNN accuracies between different convolution types supports the hypothesis presented in Section \ref{sec:errorconv}. 

In order to demonstrate the increased variance of error for grouped and depthwise convolution, all convolution outputs are extracted for the first 100 sample images of the ILSVRC2012 validation set with FP32 and Mitch-$w$6 multiplications. The errors from approximate multiplication are measured by comparing the results. The variance of accumulated error within each channel is measured as well as the variance between the convolution outputs.
The geometric means are taken across all channels as channels had wildly varying ranges of values.
Table \ref{tab:varierr} shows the measured values for various CNNs and it demonstrates the increased variance of accumulated error for grouped and depthwise convolutions as discussed in Section \ref{subsec:grouped}.
The conventional convolution results also provide the evidence that the accumulated errors have much less variance compared to the distribution of outputs, and therefore have less impact on the functionality of feature detection.

\begin{table}
	\centering
    \scriptsize
	\caption{Measured error variance with Mitch-$w$6}
    
    \vskip 0.1in
    \renewcommand\tabcolsep{3pt}
    \renewcommand\arraystretch{1.5}
    
	\begin{tabular}{l@{\hspace{7.0pt}}lcc@{\hspace{7.0pt}}ccc}
		\toprule[0.8pt]
		\textbf{Conv. Type} &
		\multicolumn{1}{l}{\textbf{Network}} &
		\multicolumn{1}{c}{\textbf{Error Vari.}} &
		\multicolumn{1}{c}{\textbf{Output Vari.}} &
		\textbf{Pct.}
		\\
        
        \cmidrule[0.5pt](l{2pt}r{2pt}){1-5}

        \textbf{Conventional}	&
		\textbf{ResNet-50} &
	    2.31E-3 &	
		6.13E-2 &
		3.8\%
		\\
	    
	    &	
		\textbf{ResNet-101} &
	    1.69E-3 &
		3.52E-2 &
	    4.8\%
		\\
		
	    &	
		\textbf{ResNet-152} &
	    1.50E-3 &
		2.72E-2 &
	    5.5\%
		\\
	    
	    &	
		\textbf{Inception-v4} &
	    6.79E-3 &
		1.22E-1 &
	    5.6\%
		\\
	    
	    &	
		\textbf{Inception-ResNet-v2} &
	    1.18E-3 &
		1.85E-2 &
	    6.3\%
		\\
	   
	    \midrule[0.4pt] 
	    \textbf{Grouped} &	
		\textbf{ResNeXt-50-32x4d} &
	    1.50E-4 &
		1.35E-3 &
	    11.2\%
	    \\
	 
	    \midrule[0.4pt] 
	    \textbf{Depthwise} &	
		\textbf{Xception} &
	    1.81E-2 &
		8.91E-2 &
	    20.4\%
		\\
	    
	    &	
		\textbf{MobileNetV2} &
	    2.00E-2 &
		1.34E-1 &
	    14.9\%
	    &	
		\\
		
		\bottomrule[0.8pt]
	\end{tabular}%
	\label{tab:varierr}%
	\vskip -0.1in
\end{table}%

While the 100 images may seem like a small number of samples, the geometric means are actually taken across millions of convolution feature maps produced from the images.
The samples include sufficient numbers of data points to demonstrate the point.
It is extremely difficult to process the entire dataset because of the large amount of internal data generated by CNNs.
Changing the sample size had little effect on the observation and the samples likely represent the behavior of the entire set for these models.

%The measured variances in Table \ref{fig:top5} do not directly correlate to the performance of Mitch-$w$6 in Fig. \ref{fig:top5} because of different network architectures. For example, ResNeXt-50-32x4d \cite{Xie2016} has poor performance with Mitch-$w$6 but shows small error variance in Table \ref{tab:varierr}, because the measured variances do not account for the error variance between different branches in the architecture.
%From Fig. \ref{fig:top5}, it is conjectured that CNNs with fewer branches perform better with approximate multiplication.

The measured variances in Table \ref{tab:varierr} do not directly correlate to the performance of Mitch-$w$6 in Fig. \ref{fig:top5} and \ref{fig:top1} because Table \ref{tab:varierr} only shows the error variance within each channel and does not account for the error variance across channels. The approximate multiplication in ResNeXt-50-32x4d causes more degradation in the prediction accuracy because ResNeXt networks have many branches in their architectures where different amounts of error accumulate. The Inception networks have relatively shorter branches and show slightly more degradation compared to the ResNet models that have none. The theoretical principle discussed in Section \ref{subsec:impact_on_conv} agrees with this analysis, though Table \ref{tab:varierr} could not capture these differences.

When the convergence of errors diminishes for grouped and depthwise convolutions, the outcomes become statistically uncertain and each CNN model may favor different approximate multipliers depending on their error patterns. DRUM6 has a different error pattern compared to Mitch-$w$6 and it performs worse than Mitch-$w$6 on the ResNeXt50 model despite the fact that it generally produces smaller errors, as shown in Fig. \ref{fig:top5} and \ref{fig:top1}. On the contrary, DRUM6 performs very well on the Xception model and it is conjectured that the errors from DRUM6 work well with this particular pre-trained model. 

\begin{figure}[t]
\vskip 0.2in
\begin{center}
\centerline{\includegraphics[width=\columnwidth]{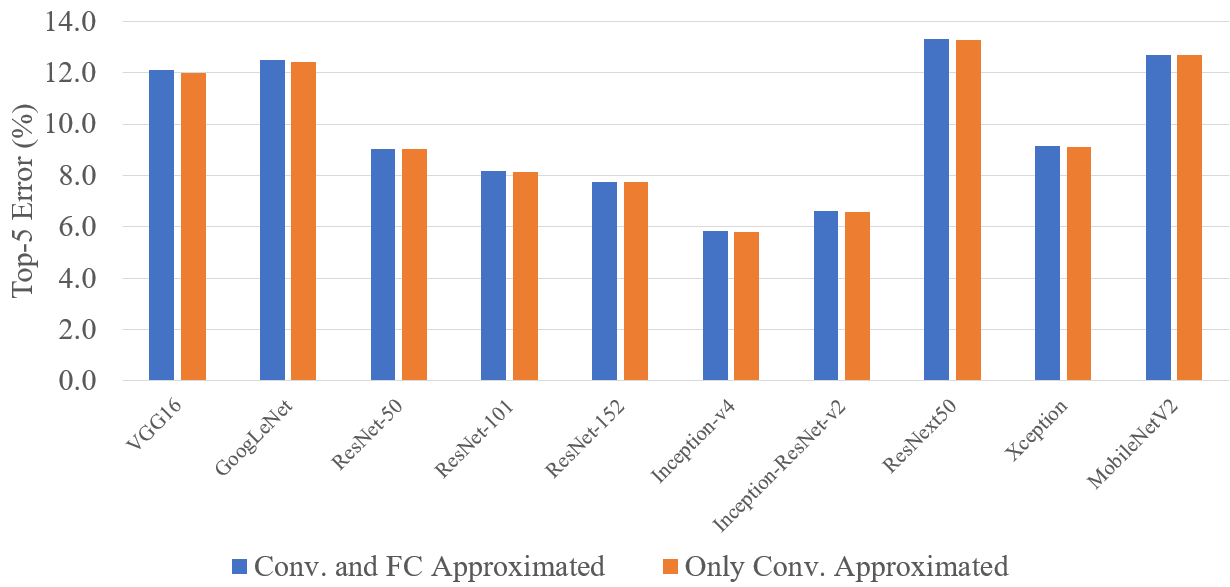}}
\caption{Low impact on CNN accuracies when FC layers do not use approximate multiplication. The experiments are performed with Mitch-$w$6.}
\label{fig:withwithoutfc}
\end{center}
\vskip -0.2in
\end{figure}

For CNNs with grouped convolutions, a sufficiently accurate approximate multiplier can still be used to perform accurate inferences, as demonstrated with the truncated iterative log multiplier in Fig. \ref{fig:top5} and \ref{fig:top1}.
When the converging effect of accumulation is reduced, the variance of accumulated error may be reduced by producing a smaller range of errors at the cost of more hardware resources.

Fig. \ref{fig:withwithoutfc} shows the effects on CNN accuracies when the FC layers perform exact multiplication instead of approximate multiplication.
Despite the fact that approximating later layers in CNNs have more influence on the outputs compared to earlier layers \cite{sarwar2016multiplier,kim2017power}, Fig. \ref{fig:withwithoutfc} demonstrates that approximating FC layers at the end of CNNs has minimal impact on CNN accuracies.
The FC layers have a large number of accumulations per each output and the higher convergence of error preserves the relative order between the final outputs.
This is the desirable property of approximate multiplication for CNN inferences as discussed in Section \ref{subsec:impact_on_conv}.

\subsection{Effect of Batch Normalization}
\label{subsec:expbat}

\begin{figure}[t]
\vskip 0.2in
\begin{center}
\centerline{\includegraphics[height=1.6in]{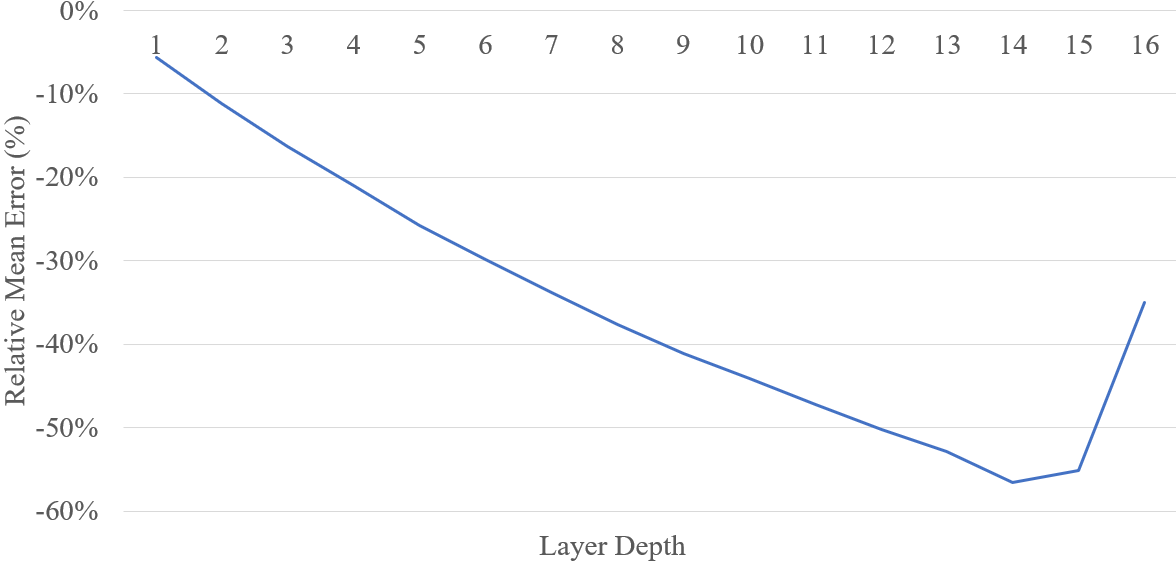}}
\caption{Accumulation of mean error on VGG16.}
\label{fig:accum}
\end{center}
\vskip -0.2in
\end{figure}

\begin{figure}[t]
\vskip 0.2in
\begin{center}
\centerline{\includegraphics[height=1.6in]{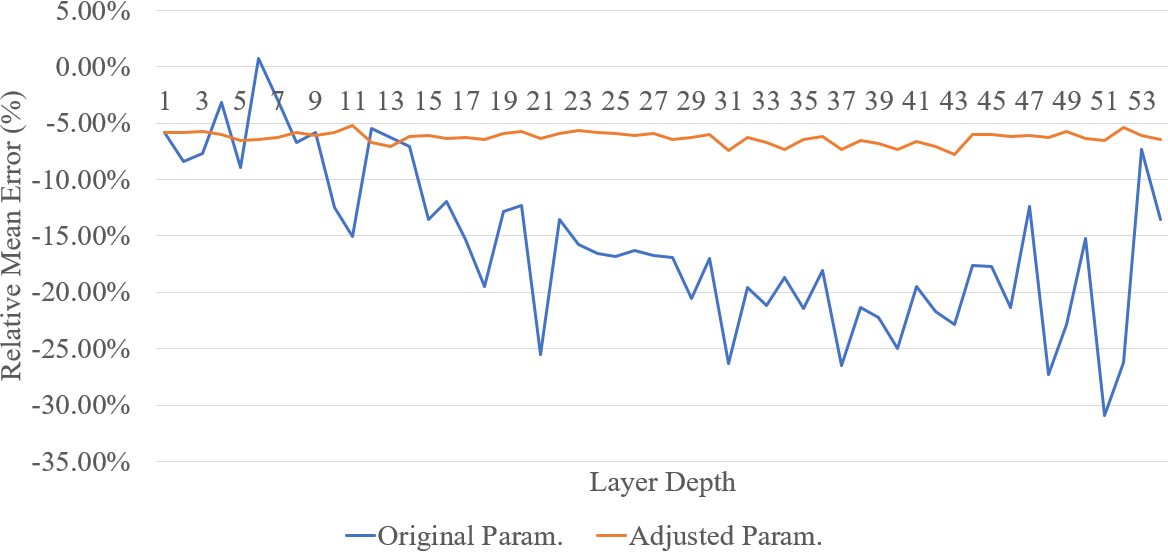}}
\caption{Effect of batch normalization on ResNet-50.}
\label{fig:resnet50}
\end{center}
\vskip -0.2in
\end{figure}

Fig. \ref{fig:accum} demonstrates the accumulation of mean error in VGG16 with Mitch-$w$6, averaged over the 100 sample images.
Because the network lacks batch normalization, the deeper layers receive the inputs that are repeatedly scaled down when the errors in multiplication are biased.
It explains the poor performance of Mitch-$w$6 on VGG16 and GoogLeNet in Fig. \ref{fig:top5}, while the unbiased DRUM6 performs well.
The last three layers that disrupt the trend are the FC layers where the added bias values become more significant when the inputs have reduced magnitudes.

Fig. \ref{fig:resnet50} shows the effect of batch normalization with properly adjusted parameters, on ResNet-50 with Mitch-$w$6 averaged over the 100 sample images. 
For Mitch-$w$6 with a mean error of -5.9\%, the mean and variance parameters in batch normalization are scaled by 0.941 and 0.885 respectively.
With the proper adjustments, batch normalization eliminates the accumulation of mean error across layers and helps approximate multiplication work with deep CNNs.
Fig. \ref{fig:resnet50} shows that the mean error per layer hovers around the mean error of Mitch-$w$6, which supports the convergence of accumulated error as well as the effectiveness of the adjusted batch normalization.
Failing to adjust the parameters not only accumulates error in deeper layers but also becomes an additional source of error with incorrect redistribution of feature maps, resulting in an unstable pattern of accumulated error.
Table \ref{tab:resnet50} shows the impact on the Top-1 and Top-5 errors of the ResNet models.
Incorrect batch normalization results in performance degradation while the corrected batch normalization layers help approximate multiplication perform well for deep ResNet models.

\begin{table}
	\centering
    \scriptsize
	\caption{Impact of batch normalization adjustment with Mitch-$w$6 on ResNet models}
   
    \vskip 0.1in 
    \renewcommand\tabcolsep{10pt}
    \renewcommand\arraystretch{1.4}
    
	\begin{tabular}{lcccccc}
		\toprule[0.8pt]
		\multicolumn{1}{r@{\hspace{8pt}}}{} &
		\multicolumn{2}{c}{\textbf{Top-1 Error}} &
		\multicolumn{2}{c}{\textbf{Top-5 Error}} 
		\\
		
		&
		\multicolumn{1}{c}{\textbf{Original}} &
		\multicolumn{1}{c}{\textbf{Adjusted}} &
		\multicolumn{1}{c}{\textbf{Original}} &
		\multicolumn{1}{c}{\textbf{Adjusted}}
		\\
        
        \cmidrule[0.5pt](l{2pt}r{2pt}){2-3}
        \cmidrule[0.5pt](l{2pt}r{2pt}){4-5}
	
		\textbf{ResNet-50} &
	    31.7\% &	
	    27.2\% &	
	    10.5\% &	
		9.0\% 
		\\
		
		\textbf{ResNet-101} &
	    31.8\% &	
	    26.0\% &	
	    12.0\% &	
		8.2\% 
		\\
		
		\textbf{ResNet-152} &
	    31.2\% &	
	    25.2\% &	
	    11.5\% &	
		7.7\% 
		\\
		
		\bottomrule[0.8pt]
	\end{tabular}%
	\label{tab:resnet50}%
	\vskip -0.1in
\end{table}%

\section{Comparison of Costs}
\label{sec:costs}

Using the bfloat16 format significantly reduces the hardware costs compared to the FP32 floating-point format and has been widely adopted in Machine Learning hardware accelerators.
While its ease of use and the ability to perform training as well as inference are undeniably advantageous, its arithmetic units are slower and consume more energy compared to the discussed multipliers based on the fixed-point format.
It is plausible to have a use-case scenario where embedded systems perform only inferences under strict design constraints, while communicating to datacenters where training occurs.
This section presents a brief comparison of the hardware costs against a bfloat16 MAC unit to give an idea of the potential benefits of the approximate log multiplication. 

Table \ref{tab:costcomp} compares the costs among the MAC units of FP32, bfloat16 and the Mitch-$w$, as synthesized with a 32nm standard library from Synopsys.
The Mitch-$w$6 HDL code is available in \cite{log_source}, the FP32 MAC design is from \cite{del2014ultra}, and we modified the FP32 design to create the bfloat16 MAC. 
Synopsys Design Compiler automatically synthesized the fixed-point MAC, and Mitch-$w$6 is followed by an exact fixed-point adder.
The 32-bit Mitch-$w$6 design represents the circuit used for the experiments while the 16-bit design represents what is potentially achievable with the proper quantization such as \cite{jacob2018quantization}.
It is clear from Table \ref{tab:costcomp} that applying approximate multiplication to CNNs can save a significant amount of resources for inferences.

The presented figures do not consider the potential benefits when adopting multiple log multipliers, where additional optimization for resource sharing can be performed depending on the design of the hardware accelerator.
Oliveira et al. \cite{oliveira2019design} proposed that certain parts of the log multiplier can be removed or shared between multiple instances of MAC units depending on the accelerator design.

\begin{table}
	\centering
    \scriptsize
	\caption{Hardware costs of FP32, bfloat16, fixed-point and Mitch-$w$6 MAC units}
    
    \vskip 0.1in 
    \renewcommand\tabcolsep{3pt}
    \renewcommand\arraystretch{1.4}
    
	\begin{tabular}{lcccccc}
		\toprule[0.8pt]
		&
		\multicolumn{3}{c}{\textbf{N=16}} &
		\multicolumn{3}{c}{\textbf{N=32}}
		\\
		
		&
		\multicolumn{1}{c}{\textbf{bfloat16}} &
		\multicolumn{1}{c}{\textbf{Fixed}} &
		\multicolumn{1}{c}{\textbf{Mitch-$w$6}} &
		\multicolumn{1}{c}{\textbf{FP32}} &
		\multicolumn{1}{c}{\textbf{Fixed}} &
		\multicolumn{1}{c}{\textbf{Mitch-$w$6}} 
		\\
        
        \cmidrule[0.5pt](l{2pt}r{2pt}){2-4}
        \cmidrule[0.5pt](l{2pt}r{2pt}){5-7}
		
		\textbf{Delay (ns)} &
		4.77 &
		2.07 &
		2.74 &
		7.52 &
		4.29 &
		4.39 
		\\
		
		\textbf{Power (mW)} &
		1.47 &
		1.17 &
		0.50 &
		5.80 &
		4.36 &
		0.98 
		\\
		
		\textbf{Energy (pJ)} &
		7.01 &
		2.42 &
		1.37 &
		43.62 &
		18.70 &
		4.30 
		\\
		
		\textbf{Energy vs. bfloat16} &
		100\% &
		35\% &
		20\% &
		622\% &
	    267\% &
		61\% 
		\\
		
		\bottomrule[0.8pt]
	\end{tabular}%
	\label{tab:costcomp}%
	\vskip -0.1in
\end{table}%

\section{Related Works}
\label{sec:related}
There have been a number of previous works that applied approximate multipliers to CNN inferences.
This paper explains the underlying reason why some of these methods perform well despite the error and how to extend the methodologies to deep CNNs with batch normalization.
To the best of our knowledge, this is the first work to demonstrate that one approximate multiplier design can perform successful inferences on the various ResNet and Inception network models without retraining.

One study in \cite{hammad2018impact} applied various approximate multipliers with varying accuracies to the VGG network, and it provided more evidence that approximate multiplication was compatible with CNN inferences.
Their work included interesting experimental results that support our hypothesis.
They found that approximating the convolution layers with higher numbers of channels resulted in less degradation of CNN accuracy, and this agrees with our finding that variance of accumulated error decreases with more inter-channel accumulations.

The works presented in \cite{du2014leveraging,mrazek2016design,ansari2019improving, mrazek2019alwann,de2018designing} had used logic minimization to create the optimal approximate multipliers for each network model. 
Logic minimization intentionally flips bits in the logic to reduce the size of the operators, and these techniques use heuristics to find the optimal targets. While these studies demonstrate promising results for improving the efficiency of CNN inferences, the heuristics involve the costly exploration of a large design space and do not ensure that the optimal multipliers for one situation would be optimal for another.
%It is difficult to create a hardware accelerator based on these approaches, because it requires to process many different instances of CNNs.

The Alphabet Set Multiplier proposed in \cite{sarwar2016multiplier} stores multiples of each multiplier value as alphabets and combines these alphabets to produce the products. Because the stored multiples require memory accesses, the authors eventually proposed the design with a single alphabet that had performed reasonably well for the simple datasets. However, the design was too inaccurate to handle the more complex dataset of ImageNet \cite{kim2018efficient}.

Approximate log multiplication from Mitchell's Algorithm had been applied to small CNN models in \cite{kim2018low, kim2018efficient, ansari2020improved}. The iterative log multipliers that increase accuracy by iterating log multiplication had been also studied \cite{lotrivc2012applicability, kim2019cost, kung2015power}.
They were mostly effective at performing CNN inferences but the reason for the good performances largely remained unsolved.
This paper provides deeper understanding of the effects of approximate multiplication on CNNs.

The log multipliers should be distinguished from the log quantization presented in \cite{lee2017lognet, miyashita2016convolutional}.
The log quantization performs all operations in the log domain and suffers from inaccurate additions, which may explain why the performances drop for more complex networks.
The Mitchell's Algorithm still performs exact additions in the fixed-point format which helps maintain the CNN performance, as discussed in Section \ref{sec:errorconv}.

There are many other ways of approximating multiplication that had not been applied to deep CNNs, such as 
\cite{liu2018design, salamat2018rnsnet, imani2018canna} among countless others. While we believe that the studied multiplier designs are the most promising, there are most likely other related opportunities for improving CNNs.

\section{Conclusion}
\label{sec:conclusion}
This paper provides a detailed explanation of why CNNs are resilient against the errors in multiplication.
Approximate multiplication favors the wide convolution layers with many input channels and batch normalization can be adjusted for deeper networks, making it a promising approach as the networks become wider and deeper to handle various real-world applications.
The proposed approximate multipliers show promising results for CNN architectures, and the arithmetic explanations provide a new and effective way for designing hardware accelerators. 
They also help explain some of the phenomenon observed in the related works while providing guidelines for extending to deeper CNNs with batch normalization.

The most widely applicable insight of this paper is that the multiplications in CNNs can be approximated while the additions have to be accurate.
The implications are far-reaching and may help analyze and justify a variety of other approximation techniques that were previously only supported by empirical evidence.
In this paper, we provide the arithmetic reason behind the success of bfloat16 approximation \cite{wang2019bfloat16} and also conjecture that log quantization \cite{miyashita2016convolutional} loses CNN accuracy because of inaccurate additions.
For quantization, the convergence theory can justify the reduced number of bits used for weights while accumulations are done with a higher number of bits.
The findings may help justify the analog processing of neural networks where the multiplication resistors may have some process variation \cite{shim2016low}.
These are only a few examples and new approximation techniques may be evaluated in the similar fashion in terms of the variance of accumulated error.
Various studies on approximation of CNN inferences have relied only on the end results as the inner workings of CNNs are often treated as black boxes.
This paper seeks to contribute towards a more analytical understanding of CNN approximation based on arithmetic.

% if have a single appendix:
%\appendix[Proof of the Zonklar Equations]
% or
%\appendix  % for no appendix heading
% do not use \section anymore after \appendix, only \section*
% is possibly needed

% use appendices with more than one appendix
% then use \section to start each appendix
% you must declare a \section before using any
% \subsection or using \label (\appendices by itself
% starts a section numbered zero.)
%

%\appendices
%\section{Proof of the First Zonklar Equation}
%Appendix one text goes here.
%
%% you can choose not to have a title for an appendix
%% if you want by leaving the argument blank
%\section{}
%Appendix two text goes here.

% use section* for acknowledgment
\ifCLASSOPTIONcompsoc
  % The Computer Society usually uses the plural form
  \section*{Acknowledgments}
\else
  % regular IEEE prefers the singular form
  \section*{Acknowledgment}
\fi
This work has been partially supported by the CPCC at UCI, the Community of Madrid under grant S2018/TCS-4423, the EU (FEDER) and the Spanish MINECO under grant RTI2018-093684-B-I00, and the NRF of South Korea funded by the Ministry of Education (2017R1D1A1B03030348).

% Can use something like this to put references on a page
% by themselves when using endfloat and the captionsoff option.
\ifCLASSOPTIONcaptionsoff
  \newpage
\fi

%\newpage

% trigger a \newpage just before the given reference
% number - used to balance the columns on the last page
% adjust value as needed - may need to be readjusted if
% the document is modified later
%\IEEEtriggeratref{8}
% The "triggered" command can be changed if desired:
%\IEEEtriggercmd{\enlargethispage{-5in}}

% references section

% can use a bibliography generated by BibTeX as a .bbl file
% BibTeX documentation can be easily obtained at:
% http://mirror.ctan.org/biblio/bibtex/contrib/doc/
% The IEEEtran BibTeX style support page is at:
% http://www.michaelshell.org/tex/ieeetran/bibtex/
\bibliographystyle{IEEEtran}
% argument is your BibTeX string definitions and bibliography database(s)
%\bibliography{IEEEabrv,../bib/paper}
\bibliography{IEEEabrv,./logNN}
%
% <OR> manually copy in the resultant .bbl file
% set second argument of \begin to the number of references
% (used to reserve space for the reference number labels box)
%\begin{thebibliography}{1}
%
%\bibitem{IEEEhowto:kopka}
%H.~Kopka and P.~W. Daly, \emph{A Guide to \LaTeX}, 3rd~ed.\hskip 1em plus
%  0.5em minus 0.4em\relax Harlow, England: Addison-Wesley, 1999.
%
%\end{thebibliography}

% biography section
% 
% If you have an EPS/PDF photo (graphicx package needed) extra braces are
% needed around the contents of the optional argument to biography to prevent
% the LaTeX parser from getting confused when it sees the complicated
% \includegraphics command within an optional argument. (You could create
% your own custom macro containing the \includegraphics command to make things
% simpler here.)
%\begin{IEEEbiography}[{\includegraphics[width=1in,height=1.25in,clip,keepaspectratio]{mshell}}]{Michael Shell}
% or if you just want to reserve a space for a photo:
%\vspace{30mm}

\begin{IEEEbiography}
[{\includegraphics[width=1in,height=1.25in,clip,keepaspectratio]{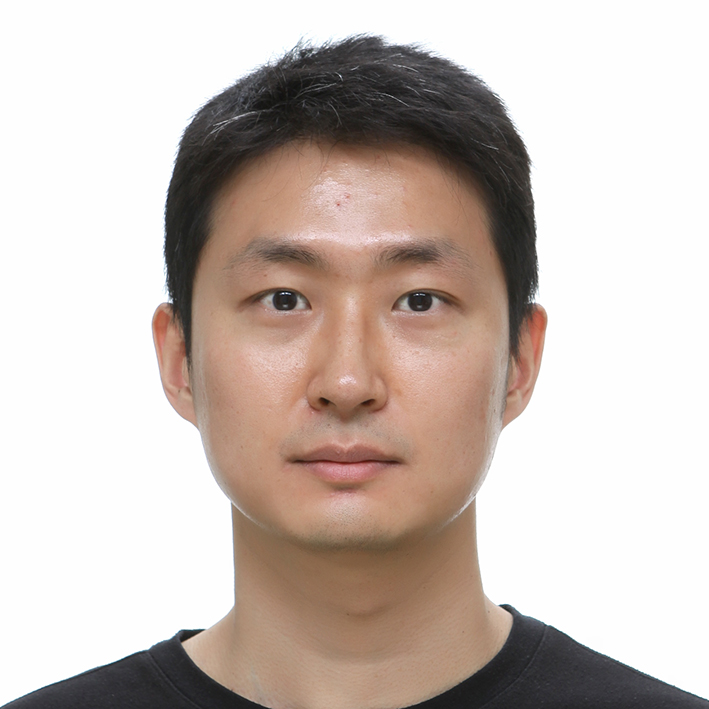}}]{Min Soo Kim}
received the BA.Sc degree in Engineering Science from the University of Toronto in 2008, and the M.S. and Ph.D. degrees in Computer Engineering from the University of California, Irvine, in 2011 and 2020 respectively. He currently works at NGD Systems as an AI Software Engineer, and his research interests include computational storage and hardware acceleration of convolutional neural networks.
\end{IEEEbiography}

\vskip 0pt plus -1fil

\begin{IEEEbiography}
[{\includegraphics[width=1in,height=1.25in,clip,keepaspectratio]{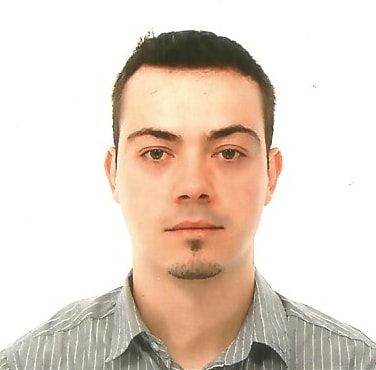}}]{Alberto A. Del Barrio}
received the Ph.D. degree in Computer Science from the Complutense University of Madrid (UCM), Madrid, Spain, in 2011. 
Since 2020, he is an Associate Professor of Computer Science with the Department of Computer Architecture and System Engineering, UCM. His research interests include Design Automation, Arithmetic as well as Video Coding Optimizations.
\end{IEEEbiography}

\vskip 0pt plus -1fil

\begin{IEEEbiography}
[{\includegraphics[width=1in,height=1.25in,clip,keepaspectratio]{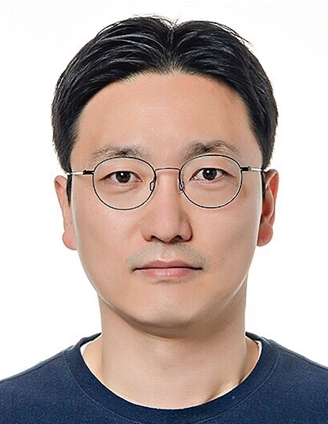}}]{HyunJin Kim}
is an associate professor in the School of Electronics and Electrical Engineering at Dankook University, Republic of Korea. He received a Ph.D in Electronics and Electrical Engineering (2010) from Yonsei University. He worked as a mixed-signal VLSI circuit designer at Samsung Electromechanics (2002$\sim$2005), and as a senior engineer in a flash memory controller project at Samsung Electronics (2010$\sim$2011). His current research interests include approximate \& stochastic computing for neural network implementation methodology, string matching engines, and energy-aware embedded systems.
\end{IEEEbiography}

\vskip 0pt plus -1fil

\begin{IEEEbiography}
[{\includegraphics[width=1in,height=1.25in,clip,keepaspectratio]{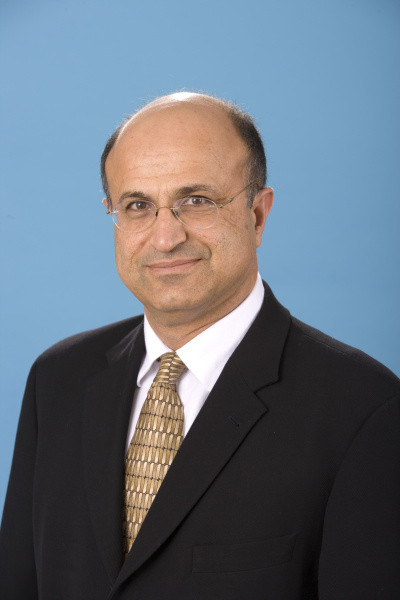}}]{Nader Bagherzadeh} is a
professor of computer engineering in the Department
of Electrical Engineering and Computer
Science at the University of California, Irvine,
where he served as a chair from 1998 to 2003.
Dr. Bagherzadeh has been involved in research
and development in the areas of: computer architecture,
reconfigurable computing, VLSI chip
design, network-on-chip, 3D chips, sensor networks,
computer graphics, memory and embedded
systems, since he received a Ph.D. degree
from the University of Texas at Austin in 1987. He is a Fellow of the IEEE.
\end{IEEEbiography}

% if you will not have a photo at all:
% \begin{IEEEbiographynophoto}{John Doe}
% Biography text here.
% \end{IEEEbiographynophoto}

% insert where needed to balance the two columns on the last page with
% biographies
%\newpage

% \begin{IEEEbiographynophoto}{Jane Doe}
% Biography text here.
% \end{IEEEbiographynophoto}

% You can push biographies down or up by placing
% a \vfill before or after them. The appropriate
% use of \vfill depends on what kind of text is
% on the last page and whether or not the columns
% are being equalized.

%\vfill

% Can be used to pull up biographies so that the bottom of the last one
% is flush with the other column.
%\enlargethispage{-5in}

% that's all folks
\end{document}